\documentclass{article}

\usepackage{amsmath}
\usepackage{amsfonts}
\usepackage{graphicx}
\usepackage{adjustbox}
\usepackage{url}
\usepackage{multirow}
\usepackage{float}
\usepackage{array}
\usepackage[normalem]{ulem}
\usepackage[justification=centering]{caption}
\usepackage[ruled,linesnumbered]{algorithm2e}
\usepackage{algpseudocode}
\usepackage[numbers]{natbib}
\usepackage{booktabs}
\usepackage[normalem]{ulem}
\usepackage{color}
\usepackage{authblk}

\newcommand{\R}{\mathbb{R}}
\newcommand{\tabitem}{~~\llap{\textbullet}~~}

\title{Task-specific Word Identification from Short Texts Using a Convolutional Neural Network}
\author[1]{Shuhan Yuan \thanks{This research was conducted while Shuhan Yuan visited University of Arkansas.}}
\author[2]{Xintao Wu}         			
\author[1]{Yang Xiang}         			
\affil[1]{Tongji University, Shanghai, China \protect\\ Email:\{4e66,shxiangyang\}@tongji.edu.cn}
\affil[2]{University of Arkansas, Fayetteville, AR, USA \protect\\ Email:xintaowu@uark.edu}
\date{\vspace{-5ex}}

\begin{document}

\maketitle

\begin{abstract}
Task-specific word identification aims to choose the task-related words that best describe a short text. Existing approaches require well-defined seed words or lexical dictionaries (e.g., WordNet), which are often unavailable for many applications such as social discrimination detection and fake review detection. However, we often have a set of labeled short texts where each short text has a task-related class label, e.g., discriminatory or non-discriminatory, specified by users or learned by classification algorithms. In this paper, we focus on identifying task-specific words and phrases from short texts by exploiting their class labels rather than using seed words or lexical dictionaries. We consider the task-specific word and phrase identification as feature learning. We train a convolutional neural network over a set of labeled texts and use score vectors to localize the task-specific words and phrases. Experimental results on sentiment word identification show that our approach significantly outperforms existing methods. We further conduct two case studies to show the effectiveness of our approach. One case study on a crawled tweets dataset demonstrates that our approach can successfully capture the discrimination-related words/phrases. The other case study on fake review detection shows that our approach can identify the fake-review words/phrases.

\end{abstract}

\newpage

\section{Introduction}

Identifying task-specific words from a short text (e.g., tweet) aims to select the task-related words that best describe the short text. Task-specific word identification has received much attention in sentiment analysis research \cite{Deng2016Identifying,Liang2014Conr,Hassan2010Identifying}. The sentiment words, which express a positive or negative polarity, are key information for text summarization and text filtering. However, these approaches often need to import either seed words or lexical dictionaries (e.g., WordNet) to identify the polarity words \cite{Deng2016Identifying,Hassan2010Identifying}.

In many applications (e.g., examining whether a tweet is discriminatory or contains private information), we do not have well-defined seed words or lexical dictionaries. Instead, we often have a set of labeled short texts each of which has a task-related class label, e.g., discriminatory/non-discriminatory or private/non-private, specified by users or learned by classification algorithms.  In this work, we focus on identifying task-specific words and phrases from short texts by exploiting their class labels rather than using seed words or lexical dictionaries. For example, given a tweet ``\textit{Do you need any more proof the blacks are just low life Savages?}''  and its label ``\textit{discriminatory}'', we aim to identify the discrimination-related words such as ``\textit{blacks}'' and ``\textit{savages}'' from this tweet text.

Task-specific word identification is important for other text analysis tasks in natural language processing and information retrieval, such as text summary and lexical dictionary construction. To understand why a short text is related to its class, it is imperative to identify and highlight its task-specific words. The identified words can be considered as the discriminative features to separate the short text from ones in other classes. Moreover, the identified task-specific words from a large text corpus can be used to build dictionaries for many important and challenging tasks on social media analysis such as detecting racism-related discrimination tweets.

In this paper, we consider the task-specific word identification as feature learning \cite{Bengio2013Representation} and use convolutional neural networks (CNNs) \cite{Lecun1998GradientBased} to identify task-specific words from a set of labeled short texts. CNNs have achieved impressive performance in different feature learning problems in computer vision, speech recognition, and text analysis areas \cite{Deng2013New,Hamid2013Exploring,Krizhevsky2012Imagenet,Kim2014Convolutional,Kalchbrenner2014Convolutional,Yin2015Convolutional}. Using CNNs to localize the positions of objects in an image based on the image-level labels has also been investigated \cite{Oquab2015Is,Zhou2016Learning}. Our CNN-based approach for task-specific word identification is similar in principle to the image object localization task because both aim to find the most discriminative features from an input. However, the image object localization methods could not be directly adapted to task-specific word identification due to two reasons.  First, the image object localization uses the raw data of an image as input to the CNN model whereas we cannot simply represent text as matrix. Second, the image object localization task is to highlight  sub-regions of the input matrix, which correspond to physical objects in an image. So even if we derive the matrix representation of text, sub-regions identified by the object localization do not correspond to words or phrases.

To address the first issue, the words in a short text need to be mapped to their feature space in our task. Traditionally, each word is represented as an one-hot vector which does not capture the semantic information about the word. Recently, mapping words to a low dimensional semantic vector space (called word embedding) is widely used as the representation of words \cite{Collobert2011Natural,Mikolov2013Efficient,Pennington2014Glove}. We adopt word embeddings to represent the words in our task. Thus, we can construct a short text matrix by combining the word embeddings. For the second problem, simply selecting sub-dimensions of the text matrix is meaningless because sub-dimensions of word embeddings are generally unexplainable. We propose an approach to determine a hidden score vector that can quantify the importance or relevancy of words to a specific task.  The words with high score values in the score vector are then considered as task-specific words.

Our task-specific word identification approach is based on training a CNN over a set of labeled texts. We show how to derive the score vectors using the parameters of the CNN model built from the set of labeled texts.  We further extend our task-specific word identification approach to task-specific phrase identification. We conduct three experiments to evaluate our approach, sentiment word identification, discrimination-related word/phrase identification, and fake review word/phrase identification. The first experiment shows that our approach can more accurately identify sentiment words than existing approaches by comparing the identified words with the ground truth. The second experiment demonstrates our approach can successfully capture meaningful discrimination-related words/phrases from a crawled tweet dataset. The third experiment shows our approach can figure out the strong polarity words which are frequently in fake reviews.

The rest of this paper is organized as follows. In section \ref{sec:rw}, we first briefly review of  the related work on feature selection and sentiment word identification, along with research on deep neural networks for short text modeling. We then introduce our model for task-specific word identification, which computes a score vector for evaluating the weights of words in a short text by using a well-trained convolutional neural network. To evaluate our model, we conduct three experiments, which are sentiment word identification, discrimination-related word/phrase identification and fake review word/phrase identification. The experimental results show the effectiveness of our model. Finally, we conclude our work.

\section{Related Work}
\label{sec:rw}

\subsection{Feature selection and sentiment word identification}

The feature selection methods for text classification can adapt to identify the task-specific words. The feature selection can reduce the dimension of input space and identify the discriminative features to improve the performance of classification. There are a large number of unsupervised feature selection methods based on statistical measures, like term frequency, information gain, mutual information, and term strength \cite{Dasgupta2007Feature,Yang1997Comparative,Makrehchi2017Extracting,Uysal2016Improved}. Some other methods are based on dimensionality reduction including principle component analysis (PCA), linear discriminant analysis, and locally linear embedding \cite{Li2016Feature}. However, feature selection as a preprocessing step of data mining and machine learning applications is used for classification or clustering. Our method combines the label information of the text to identify the discriminative features, which can further improve the performance of selecting task-specific words.

Sentiment word identification, as a special case of task-specific word identification, usually requires seed words. Some of the approaches are based on the similarity between the words and seed words. For example, \cite{Turney2003Measuring,Qiu2009Expanding} adopt point-wise mutual information to measure the similarity. \cite{Kanayama2006Fully} assumes the same polarities would appear successively of the seed words in contexts and defines the context coherent to measure the similarity. \cite{Jijkoun2010Generating} aims to generate topic-specific subjectivity lexicons from a general polarity lexicon using a bootstrapping method. However, the seed words are selected manually and domain-dependence, which are usually incomplete. Heavy relying on seed words restricts the performance of identifying the task-specific words. Recently, some other methods \cite{Yu2013Identifying,Liang2014Conr,Park2015Efficient} are proposed to identify sentiment words based on optimization models without using seed words. \cite{Lu2011Automatic} focuses on identifying the sentiment lexicons which have different meanings in different aspects using an optimization framework. However, these methods are limited to identify sentiment words from documents and are not suitable for short texts. Meanwhile, topic models such as Latent Dirichlet allocation (LDA) \cite{Blei2003Latent} can identify topic words from a corpus as task-specific words. \cite{Zheng2014Incorporating} further incorporate appraisal expression patterns to LDA for aspect and sentiment word identification. However, LDA usually requires a large number of documents and is not suitable for short texts either \cite{Tang2014Understanding}.

Closely related to our approach is weakly supervised class saliency maps \cite{Simonyan2013Deep,Li2015Visualizing} which are widely used in computer vision area for object localization and class saliency visualization. However, as shown in our experiment evaluation, the performance of class saliency maps on task-specific words is poor. In this work, we propose a weakly supervised method to identify task-specific words from short texts based on the convolutional neural network. Our method is trained on labeled text corpus, so no seed words or other additional annotation are required. Meanwhile, our method further combines the label information to identify the task-specific words. Thus, our method is suitable for applications such as social discrimination detection from texts and fake review detection.

\subsection{Deep neural networks for short text modeling}

Deep neural networks have achieved promising results in natural language processing, like text classification \cite{Kim2014Convolutional}, question answering \cite{Weston2015Towards,Bordes2014Question}, and machine reading \cite{Hermann2015Teaching,Cheng2016Long}.

The fundamental of applying the deep neural networks for natural language processing is word embeddings \cite{Mikolov2013Efficient,Bengio2003Neural} which map each word to a dense vector. These word embeddings are trained in an unsupervised way on a large text corpus. Word embeddings can avoid the use of hand-designed features and capture the hidden semantic and grammatical features of words. Thus, word embeddings can improve the performance of many natural language processing tasks \cite{Collobert2011Natural}. To further compose the representations of phrases and short texts, the idea of semantic composition is applied to the word embeddings. The basic model is based on the algebraic operations, like additive and multiplication, to build the short text vector from word embeddings \cite{Mitchell2010Composition}. However, simple algebraic operations cannot capture the complicated structure of the natural language. Some complex models based on deep neural networks are proposed recent years. The recursive neural network \cite{Socher2013Recursive} can construct the grammar tree-like structure to represent the short text in order to capture the grammar information. The recurrent neural network \cite{Graves2013Generating} processes the short text word by word in order, which capture the sequential information of a short text. Researchers also proposed a tree-structured recurrent neural network \cite{Tai2015Improved} to combine the advantages of recursive neural network and recurrent neural network.

In this work, we adopt the convolutional neural network which as a primary model of deep neural networks has also achieved great performance on different areas, like computer vision \cite{Krizhevsky2012Imagenet}, speech recognition \cite{Hamid2013Exploring} and natural language processing \cite{Kim2014Convolutional}. Because of somewhat opaque of CNN, researchers try to demystify and understand why the performance of CNN is promising. In computer vision area, saliency maps \cite{Simonyan2013Deep,Li2015Visualizing} and deconvolution networks \cite{Zeiler2011Adaptive,Zeiler2014Visualizing} are proposed to explain the model. However, there is little work to explain the internal operation and behavior of CNN model working on the text. Our work can provide some insights about the CNN on text classification.

\section{Model Description}
\label{sec:model}

In this section, we describe our approach for identifying task-specific words or phrases. Our approach first trains the CNN model and transfers the well-learned representation of sentence to identify the task-specific words and phrases. We first introduce how to construct score vectors to identify the task-specific words based on the convolutional neural network. Then, we extend our approach to identify the task-specific phrases.

\subsection{Problem Statement}
\label{sec:ps}

Given a corpus of labeled short texts $\mathcal{X}$, each short text contains $n$ words and has the class label $c$, which can be described as $X=[x_1,x_2,\dots,x_n]$ where $x_i$ denotes the $i$-th word and the class label $c$ is from a set $C$. We assume that there is a hidden score vector $\mathbf{s}_c=[s_1,s_2,\dots,s_n]$ for the short text $X$ where $s_i$ measures the importance or relevancy of the word $x_i$ to the class $c$. Similarly, there is a score vector to measure the importance of corresponding phrases for class $c$. Thus, our task is to derive the score vector $\mathbf{s}_c$ based on the text $X$ and its label $c$ to locate  task specific words or phrases. Important notations are summarized in Table \ref{tb:notation}.

\begin{table}[]
\centering
\caption{Notation Table}
\label{tb:notation}
\begin{tabular}{ll}
\hline
Notation                  & Description                                                                                                                           \\ \hline \hline
$X=[x_1,x_2,\dots,x_n]$   & a short text containing $n$ words                                                                                                       \\ \hline
$\mathbf{x}_i \in \R^d$   & a $d$-dimensional word embedding of the $i$-th word                                                                                   \\ \hline
$\mathbf{W} \in \R^{h*d}$ & \begin{tabular}[c]{@{}l@{}}a \textit{filter} in a convolution operation\\  applied to $h$ continuous word embeddings\end{tabular}   \\ \hline
$\mathbf{s}_c$            & \begin{tabular}[c]{@{}l@{}}a score vector that indicates the importance\\  scores of words in a given text for class $c$\end{tabular} \\ \hline
\end{tabular}
\end{table}

\subsection{Convolutional Neural Network for short text Representation}
The short text representation by deep learning models first uses the word embeddings to represent the words in the short text and then performs a semantic composition over the word embeddings to build the short text representation. In this work, our task-specific word identification approach is based on the convolutional neural network first introduced in \cite{Kim2014Convolutional} to compose the short text representation. In this section, we first give a brief review about the CNN model for short text representation.

Using CNN to model the short text representation, we first map each word $x_i$ in the text $X$ to a $d$-dimensional real-valued vector space $\mathbf{x}_i \in \R^d$. These word embeddings are trained in an unsupervised way on a large text corpus. Word embeddings can avoid the use of hand-designed features and capture the hidden semantic and grammatical features of words. An embedding matrix $\mathbf{X} = [\mathbf{x}_1,\mathbf{x}_2,\dots,\mathbf{x}_n]$ is then constructed by combining each word embedding of the text, where $\mathbf{X}\in \R^{n*d}$.

Then, a convolution operation which involves a \textit{filter} $\mathbf{W} \in \R^{h*d}$ is applied to $h$ continuous word embeddings $\mathbf{X}_{j:j+h-1}$ to generate a hidden feature of these $h$ words:
\begin{equation}
\label{eq:filter}
v_j = g(\mathbf{W} * \mathbf{x}_{j:j+h-1} + b),
\end{equation}
where $b$ is the bias; $*$ is a two-dimensional convolution operation and $g$ indicates a non-linear function. The filter slides through the whole short text matrix $\mathbf{X}$ and generates a feature vector $\mathbf{v}=[v_1,v_2,\dots,v_{n-h+1}]$. After that, a pooling operation is applied to the feature vector $\mathbf{v}$. There are two widely used pooling operation: max pooling and average pooling. The max pooling operation aims to capture the most superior part of the feature vector by keeping its highest value and is defined as:
\begin{equation}
\label{eq:max}
r_{max} = max(v_1,v_2,...,v_{n-h+1}).
\end{equation}
The average pooling aims to capture all discriminative features of a text and is defined as:
\begin{equation}
\label{eq:avg}
r_{avg} = \frac{1}{n-h+1} \sum_{j=1}^{n-h+1} v_j.
\end{equation}

To capture different aspects of features from the text, the model usually applies multiple filters with different sizes of windows $h$ to generate the feature vector. After applying the pooling operation on each feature vector, the model encodes the input short text to a representation vector $\mathbf{r}=[r_1,r_2,\cdots,r_m]$, where $m$ is the number of feature vectors generated by $m$ different filters. A softmax classifier is applied on top of the representation vector $\mathbf{r}$ to predict the labels of the short texts. The softmax function is defined as:
\begin{equation}
p(y=c|\mathbf{r};\mathbf{U})=\frac{exp(\mathbf{u}_c^T \mathbf{r})}{\sum_{c' \in C} exp(\mathbf{u}_{c'}^T \mathbf{r})},
\end{equation}
where $\mathbf{U}$ is the parameters of the softmax function and $\mathbf{u}_c$ is the $c$-th column of $\mathbf{U}$; and $c$ is the class of the given text.
We adopt the cross-entropy loss function to train the CNN model:
\begin{equation}
\label{eq:loss}
L(y) = - \log P(y=c|\mathbf{r};\mathbf{U}).
\end{equation}
The parameters of the model are updated by the backpropagation algorithm.

\begin{figure*}
	\centering
	\includegraphics[width=0.9\textwidth,keepaspectratio]{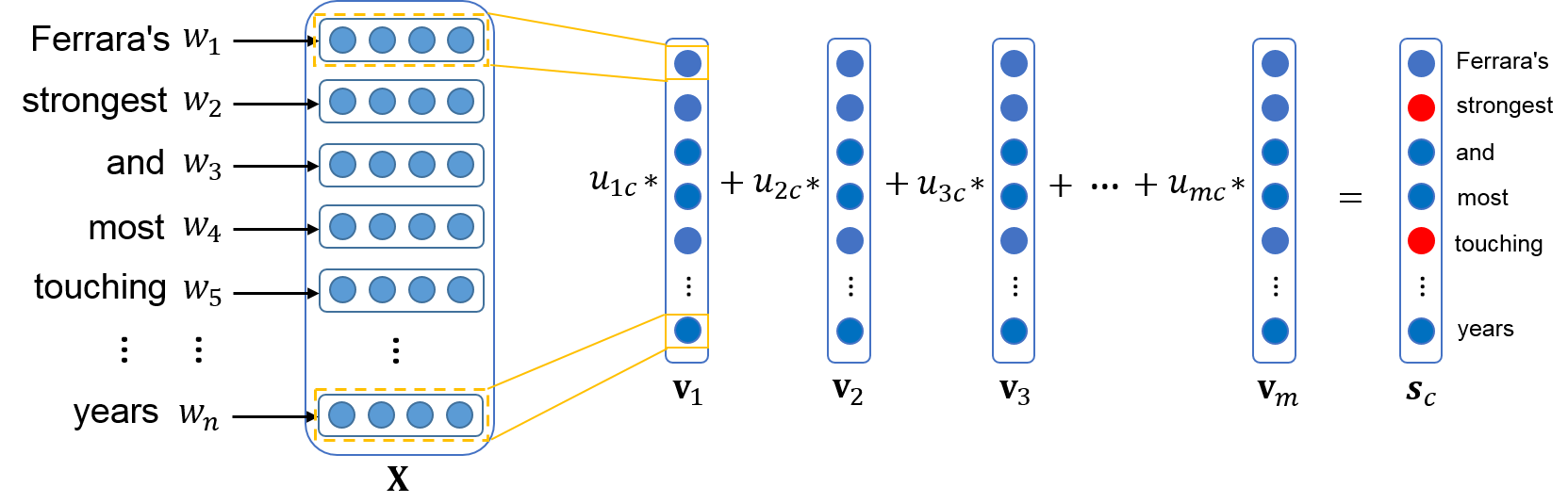}
	\caption{Given a text: \textit{Ferrara's strongest and most touching movie of recent years}, the score vector $\mathbf{s}_c$ locates the task-specific words.}
	\label{fig:sv}
\end{figure*}

\subsection{Task-specific Word Identification}
\label{sec:twi}
To identify the task-specific words, we adopt a CNN model to learn the score vector $\mathbf{s}_c$ based on the text $X=[x_1,x_2,\dots,x_n]$ and its label $c$. The text words having the Top-k highest values in the score vector $\mathbf{s}_c$ are then highlighted as task-specific words.

Given a short text $X=[x_1,x_2,\dots,x_n]$ with $n$ words, we first map the short text to a short text matrix $\mathbf{X} = [\mathbf{x}_1,\mathbf{x}_2,\dots,\mathbf{x}_n]$, where $\mathbf{X}\in \R^{n*d}$ and $\mathbf{x}_n$ is the corresponding word embedding of the word $x_n$. We apply the one word \textit{filter} $\mathbf{w} \in \R^{d}$ to each word embedding in the matrix $\mathbf{X}$ to build a convolutional layer. For word $x_j$, we learn its feature $v_j$ by:
\begin{equation}
\label{eq:conv}
v_j = g(\mathbf{w} * \mathbf{x}_{j} + b),
\end{equation}
 where $b$ is the bias; $*$ indicates the one-dimensional convolution operation and $g$ is a non-linear function. The feature $v_j$ is trained to represent a linguistic feature of the word $x_j$ in $X$. Then, the filter passes through the whole matrix $\mathbf{X}$  and produces a feature vector:
\begin{equation}
\label{eq:fv}
\mathbf{v}=[v_1,v_2,\dots,v_{n}].
\end{equation}

After obtaining the feature vector $\mathbf{v}$, a pooling operation is applied. The pooling operation is able to capture the salient information of the feature vector $\mathbf{v}$. In our model, we adopt both the max pooling and average pooling operation to extract the feature from the feature vector as the feature extracting from one filter. The max pooling operation extracts the maximum value $r_{max}=max(\mathbf{v})$ of the feature vector $\mathbf{v}$:
\begin{equation}
r_{max} = max(v_1,v_2,...,v_{n}).
\end{equation}
The average pooling operation computes the mean value $r_{avg}=mean(\mathbf{v})$ of the feature vector $\mathbf{v}$:
\begin{equation}
r_{avg} = \frac{1}{n} \sum_{j=1}^{n} v_j.
\end{equation}

Note that one feature vector $\mathbf{v}$ is computed by the same filter $\mathbf{w}$ in a CNN. Thus, the feature vector $\mathbf{v}$ actually captures one hidden semantic meaning of the input text. For modeling the short text, we need more feature vectors to capture the different aspects of hidden semantic meanings. In our work, we derive $m$ feature vectors, $\mathbf{v}_1, \cdots, \mathbf{v}_m$.  Each feature vector $\mathbf{v}_i$ is computed by a unique $\mathbf{w}_i$ and $b_i$. After applying the pooling operation on each feature vector $\mathbf{v}_i$, we get its corresponding feature $r_i$  and then combine them in one vector $\mathbf{r}$:
\begin{equation}
\mathbf{r}=[r_1,r_2,\cdots,r_m].
\end{equation}

In order to learn the representation vector $\mathbf{r}$ of $X$, we train the CNN model in a supervised way to predict the class of text $X$ by using a softmax function:
\begin{equation}
\label{eq:softmax}
p(y=c|\mathbf{r};\mathbf{U})=\frac{exp(\mathbf{u}_c^T \mathbf{r})}{\sum_{c' \in C} exp(\mathbf{u}_{c'}^T \mathbf{r})},
\end{equation}
where $\mathbf{U}$ is the parameters of the softmax function and $\mathbf{u}_c$ is the $c$-th column of $\mathbf{U}$; and $c$ is the class of the given text. We adopt the cross entropy loss function defined in Equation \ref{eq:loss} to train the model. After training the CNN for classification, the representation vector $\mathbf{r}$ expects to encode the semantic information about the input text and its class.

Next we show how we derive the score vector $\mathbf{s}_c=[s_1,s_2,\dots,s_n]$ by using parameters $\mathbf{U}$ and feature vectors $\mathbf{v}_1, \cdots, \mathbf{v}_m$, all of which are well trained in the CNN.
From the softmax function shown in Equation \ref{eq:softmax}, we can notice that $\mathbf{u}_c^T \mathbf{r} = \sum_i {u_{ic} r_i }$, where $u_{ic}$ is the entry in the $i$-th row and the $c$-th column of softmax parameter $\mathbf{U}$. Therefore, we can consider $u_{ic}$ as the weight corresponding to feature $r_i$ for class $c$. This also indicates that $u_{ic}$ captures the importance of each feature vector $\mathbf{v}_i$ for predicting class $c$, because $r_i$ is computed by the pooling operation on $\mathbf{v}_i$. Hence, the score vector $\mathbf{s}_c$ is computed by projecting the weights of softmax on to feature vectors $\mathbf{v}_i$:
\begin{equation}
\label{eq:sv}
\mathbf{s}_c = \sum_{i=1}^m u_{ic} \mathbf{v}_i.
\end{equation}

We illustrate the procedure of generating score vectors using an example in Figure \ref{fig:sv}.
Recall that each value in $\mathbf{v}_i$ measures one aspect importance of a word to a text $X$, and its weight $u_{ic}$ measures the importance to the class $c$. Thus, based on Equation \ref{eq:sv}, the values in $\mathbf{s}_c$ indicate the importance scores of words in a given text for class $c$. For example, if the $i$-th value is the highest value in the score vector $\mathbf{s}_c$, the corresponding word $x_i$ contains the most important information about class $c$. Then, we select Top-k words which have the highest score values in score vector $\mathbf{s}_c$ as the task-specific words. Our approach also provides a way to understand the internal behavior of the CNN model. The words with the corresponding higher values in the score vector $\mathbf{s}_c$ are the key information for the CNN model to predict the label of a sentence.

We show pseudo code of our approach in Algorithm \ref{algr:cnn}. For the task-specific word identification, we set the filter size $h=1$. We train the CNN model in Lines 2-17 and use the trained CNN model to identify task-specific words in Lines 18-22. For each  text $X$ in the training data $\mathcal{X}$, we first compose its text matrix $\mathbf{X}$ using word embeddings in Line 4. We generate feature vector $\mathbf{v}$ by applying the \textit{filter} $\mathbf{W} \in \R^{h*d}$ on the text matrix $\mathbf{X}$ based on Eq. \ref{eq:filter} in Line 7. We then get  hidden feature value $r$ by applying the pooling operation on $\mathbf{v}$ by Eq. \ref{eq:max} for max pooling or Eq. \ref{eq:avg} for average pooling in Line 8. After applying multiple filters, we construct the short text representation vector $\mathbf{r}=[r_1,r_2,\cdots,r_m]$ in Line 11 and predict text label by using the softmax function $p(y=c|\mathbf{r};\mathbf{U})$ by Eq. \ref{eq:softmax} in Line 12. Finally we compute the loss function $L(y)$ by Eq. \ref{eq:loss} in Line 13 and update parameters of CNN by the backpropagation algorithm in Line 14. We then use the trained CNN model to generate feature vectors of text $X$ by Eq. \ref{eq:fv} in Line 19. We apply Eq. \ref{eq:sv} to generate score vector $\mathbf{s}_c$ in Line 20 and derive  task-specific words for text $X$ which have the Top-k highest values in the score vector $\mathbf{s}_c$ in Line 21.
	
\begin{algorithm}[]
	\DontPrintSemicolon
	\SetKwInOut{Inputs}{Inputs}\SetKwInOut{Outputs}{Outputs}
	\Inputs{Training dataset $\mathcal{X}$, maximum training epoch $Epoch$, number of filters $m$, filter size $h$}
	\Outputs{Task-specific words for each text $X$ in $\mathcal{X}$}
	$k \leftarrow 0$;

	\While{$k<Epoch$}{
		\For {each short text $X$ in $\mathcal{X}$}{
			Compose the text matrix $\mathbf{X}$ for $X$ using word embeddings;

            $l \leftarrow 0$;

			\For {$l<m$}{

				Generate feature vector $\mathbf{v}$ by applying the \textit{filter} $\mathbf{W} \in \R^{h*d}$ on the text matrix $\mathbf{X}$ based on Eq. \ref{eq:filter};
				\label{line:filter}

				Apply the pooling operation on $\mathbf{v}$ by Eq. \ref{eq:max} for max pooling or Eq. \ref{eq:avg} for average pooling to get hidden feature value $r$;

				$l \leftarrow l+1$;
			}

			Construct the short text representation vector $\mathbf{r}=[r_1,r_2,\cdots,r_m]$;

			Compute the softmax function $p(y=c|\mathbf{r};\mathbf{U})$ by Eq. \ref{eq:softmax};

			Compute the loss function $L(y)$ by Eq. \ref{eq:loss};

			Update parameters of CNN by the backpropagation algorithm;
		}

		$k \leftarrow k+1$;
	}
	\For {each short text $X$ in $\mathcal{X}$}{

		Use the trained CNN model to generate feature vectors of text $X$ by Eq. \ref{eq:fv};

		Generate score vector $\mathbf{s}_c$ by Eq. \ref{eq:sv};

		Get task-specific words/phrases for text $X$ which have the Top-k highest values in the score vector $\mathbf{s}_c$
	}
	\Return the task-specific words/phrases for each text $X$ in $\mathcal{X}$

\caption{Task-specific Word/Phrase Identification}
\label{algr:cnn}
\end{algorithm}

\subsection{Task-specific Phrase Identification}
\label{sec:twp}

A phrase is defined as a concatenation of $h$ continuous words ($h$-grams). For the text $X=[x_1,x_2,\dots,x_n]$, we have its phrases $P_X=[p_1, p_2, \cdots, p_{n-h+1}]$, where $p_j$ denotes the concatenation of words $x_j, x_{j+1},\cdots, x_{j+h-1}$. Thus, the $j$-th phrase embedding is $\mathbf{X}_{j:j+h-1}=[\mathbf{x}_j,\mathbf{x}_{j+1},\cdots,\mathbf{x}_{j+h-1}]$. To identify task-specific phrases from each text $X$, we adopt the same idea of deriving a score vector $\mathbf{s}_c=[s_1,s_2,\dots,s_{n-h+1}]$ to measure the importance of each phrase in the text.

We set a phrase filter $\mathbf{W} \in \R^{h*d}$ to cover a window of $h$ words. Thus, in the CNN model, each feature $v_j$ captures the hidden feature of phrase $p_j$ about class $c$ by applying the filter $\mathbf{W}$ to $\mathbf{X}$:
\begin{equation}
\label{eq:conv_p}
v_j = g(\mathbf{W} * \mathbf{X}_{j:j+h-1} + b),
\end{equation}
where $b$ is the bias; $*$ is the two-dimensional convolution operation and $g$ indicates a non-linear function. Then, the feature vector is $ \mathbf{v}=[v_1,v_2,\dots,v_{n-h+1}]$.
The max pooling operation is defined as:
\begin{equation}
r_{max} = max(v_1,v_2,...,v_{n-h+1}).
\end{equation}
The average pooling is defined as:
\begin{equation}
r_{avg} = \frac{1}{n-h+1} \sum_{j=1}^{n-h+1} v_j.
\end{equation}

We follow the same procedure described in Algorithm \ref{algr:cnn} with the filter size $h$ to first train the CNN model for text classification, produce the score vector $\mathbf{s}_c$, and output task-specific phrases with length $h$.

\section{Experiments}

To evaluate the effectiveness of our approach, we conduct three experiments. In the first experiment, we focus on identifying sentiment words using two publicly available datasets with the ground truth.
In the second experiment, we conduct a case study of identifying social discrimination-related words or phrases from tweets. We crawl our own datasets and focus on two types of social discriminations, sexism and racism. In the third experiment, we conduct another case study of identifying fake review words or phrases based on a fake review dataset.

{\noindent \bf Word Embeddings and Hyperparameters.}
We use the off-the-shelf pre-trained word embeddings (\url{https://code.google.com/archive/p/word2vec/}) \cite{Mikolov2013Efficient} and randomly initialize the words that do not have pre-trained word embeddings. The dimension of word embeddings $d$ is 300. The number of feature vectors $m$ is 100.

{\noindent \bf Baselines.}
In our approach, we evaluate two pooling operations in our CNN model and name the score vectors with average pooling operation and max pooling operation \textbf{SV-AVG} and \textbf{SV-MAX}, respectively. We compare our approach with the following baselines for task-specific word identification.
\begin{itemize}
	\item TF-IDF: TF-IDF, widely used as a feature selection method in information retrieval and text mining, is a statistic to reflect how important a word is to a document in a collection or corpus.
	We calculate the TF-IDF value of words on the positive and negative text corpus separately. For each text, we output the Top-k words with the highest TF-IDF values.
	\item TF-IDF-softmax: We apply the softmax classifier on the TF-IDF features. After training the model for text classification, we follow the similar procedure described in last section to identify the task-specific words. The score vector is generated by multiplying the parameters $\mathbf{u}_c$ of softmax with TF-IDF values of each text for the class $c$.
	\item Saliency Map: Saliency Map was proposed as a CNN visualization technique \cite{Simonyan2013Deep}. It computes the gradient of the class score with respect to the input and can identify the discriminative features of the input. The authors \cite{Denil2014Modelling} applied this technique to highlight the important sentences of a document. In our experiment, we consider the words in a text with high absolute values of the derivative on word embeddings as task-specific words. The saliency map can be derived based on CNN with max pooling or CNN with average pooling. We denote the saliency map using CNN with max pooling (average pooling) as \textbf{SalMap-MAX} (\textbf{SalMap-AVG}).

\end{itemize}

{\noindent \bf Evaluation Metric.}
For sentiment word and discrimination-related word identification, we adopt accuracy@k, precision, recall, and F1 to evaluate the performance.

\begin{itemize}
	\item \textit{Accuracy@k} is a metric which calculates the fraction of the Top-k words selected by each method with the ground truth $T$.

	\item  We further evaluate our approach based on the \textit{precision},\textit{recall} and \textit{F1}. We  obtain the Top-k words of each text with highest values in its score vector and compare the selected words with the ground truth dictionary.

\end{itemize}

\begin{table}[tbp]
\centering
\caption{Accuracy@k on sentiment word identification}
\label{tb:accuracy}
\begin{tabular}{cccccc}
\hline
Dataset              & Method         & \begin{tabular}[c]{@{}c@{}}Classification\\  Accuracy\end{tabular} & Top-1            & Top-3            & Top-5            \\ \hline
\multirow{5}{*}{MR}  & TF-IDF         &N/A                                                                      & 22.78\%          & 23.19\%          & 22.87\%          \\ \cline{2-6}
                     & TF-IDF-Softmax & 73.92\%                                                            & 43.93\%          & 34.10\%          & 28.39\%          \\ \cline{2-6}
                     & SalMap-MAX     & N/A                                                                    & 29.55\%          & 25.28\%          & 21.38\%          \\ \cline{2-6}
                     & SalMap-AVG     & N/A                                                                    & 17.52\%          & 12.71\%          & 11.22\%          \\ \cline{2-6}
                     & SV-MAX         & \textbf{78.32\%}                                                   & 61.01\%          & 40.04\%          & 29.12\%          \\ \cline{2-6}
                     & SV-AVG         & 76.87\%                                                            & \textbf{66.80\%} & \textbf{43.28\%} & \textbf{30.43\%} \\ \hline
\multirow{5}{*}{SST} & TF-IDF         & N/A                                                                    & 25.14\%          & 25.08\%          & 23.55\%          \\ \cline{2-6}
                     & TF-IDF-Softmax & 75.04\%                                                            & 50.67\%          & 37.51\%          & 31.03\%          \\ \cline{2-6}
                     & SalMap-MAX     & N/A                                                                    & 35.67\%          & 29.51\%          & 23.83\%          \\ \cline{2-6}
                     & SalMap-AVG     & N/A                                                                    & 17.83\%          & 14.81\%          & 11.78\%          \\ \cline{2-6}
                     & SV-MAX         & \textbf{82.33\%}                                                   & 68.35\%          & 42.90\%          & 30.50\%          \\ \cline{2-6}
                     & SV-AVG         & 81.23\%                                                            & \textbf{71.24\%} & \textbf{45.32\%} & \textbf{31.25\%} \\ \hline
\end{tabular}
\end{table}

\begin{table}[]
\centering
\caption{Precision, recall and F1 on sentiment word identification}
\label{tb:f1}
\begin{adjustbox}{max width=0.95\textwidth}
\begin{tabular}{ccccccccccc}
\hline
\multirow{2}{*}{Dataset} & \multirow{2}{*}{Method} & \multicolumn{3}{c}{Top-1}                             & \multicolumn{3}{c}{Top-3}                             & \multicolumn{3}{c}{Top-5}                             \\ \cline{3-11}
                         &                         & Precision        & Recall           & F1               & Precision        & Recall           & F1               & Precision        & Recall           & F1               \\ \hline
\multirow{5}{*}{MR}      & TF-IDF                  & 30.48\%          & 5.12\%           & 8.77\%           & 29.85\%          & 15.05\%          & 20.01\%          & 28.57\%          & 24.00\%          & 26.08\%          \\ \cline{2-11}
                         & TF-IDF-Softmax          & 45.83\%          & 23.93\%          & 31.44\%          & 34.72\%          & 54.37\%          & 42.38\%          & 28.78\%          & 75.10\%          & 41.61\%          \\ \cline{2-11}
                         & SalMap-MAX	           & 16.75\%          & 8.32\%           & 11.11\%          & 13.01\%          & 19.40\%          & 15.56\%          & 11.32\%          & 28.12\%          & 16.13\%          \\ \cline{2-11}
                         & SalMap-AVG              & 17.52\%          & 8.66\%           & 11.59\%          & 12.71\%          & 18.85\%          & 15.18\%          & 11.22\%          & 27.72\%          & 15.98\%          \\ \cline{2-11}
                         & SV-MAX                  & 59.65\%          & 30.36\%          & 40.24\%          & 39.66\%          & 60.54\%          & 47.92\%          & 28.69\%          & 72.98\%          & 41.19\%          \\ \cline{2-11}
                         & SV-AVG                  & \textbf{66.93\%} & \textbf{33.79\%} & \textbf{44.91\%} & \textbf{43.22\%} & \textbf{65.44\%} & \textbf{52.05\%} & \textbf{30.32\%} & \textbf{76.53\%} & \textbf{43.43\%} \\ \hline \hline
\multirow{5}{*}{SST}     & TF-IDF                  & 34.22\%          & 6.46\%           & 10.87\%          & 31.47\%          & 17.82\%          & 22.75\%          & 28.44\%          & 26.74\%          & 27.62\%          \\ \cline{2-11}
                         & TF-IDF-Softmax          & 51.41\%          & 27.64\%          & 35.94\%          & 37.28\%          & 60.13\%          & 46.03\%          & 30.81\%          & \textbf{82.79\%} & 44.90\%          \\ \cline{2-11}
                         & SalMap-MAX	           & 20.00\%          & 9.16\%           & 12.13\%          & 13.79\%          & 21.28\%          & 16.85\%          & 11.68\%          & 29.66\%          & 16.75\%          \\ \cline{2-11}
                         & SalMap-AVG              & 17.83\%          & 8.66\%           & 11.66\%          & 13.81\%          & 20.59\%          & 16.57\%          & 11.78\%          & 28.62\%          & 16.69\%          \\ \cline{2-11}
                         & SV-MAX                  & 66.62\%          & 34.53\%          & 45.49\%          & 42.56\%          & 66.21\%          & 51.82\%          & 30.22\%          & 78.33\%          & 43.61\%          \\ \cline{2-11}
                         & SV-AVG                  & \textbf{71.32\%} & \textbf{36.71\%} & \textbf{48.47\%} & \textbf{45.32\%} & \textbf{70.00\%} & \textbf{55.02\%} & \textbf{31.38\%} & 80.78\%          & \textbf{45.20}   \\ \hline
\end{tabular}
\end{adjustbox}
\end{table}

\subsection{Sentiment Word Identification}

The first experiment is to identify the sentiment words from each text. We only consider the binary classification problem, so the objective of our model is to select the positive and negative sentiment words.

{\noindent \bf Datasets.}
We evaluate the performance of sentiment word identification on two datasets --- the Movie Review dataset (\textbf{MR}) \cite{Pang2005Seeing} and the Stanford Sentiment Treebank dataset (\textbf{SST}) \cite{Socher2013Parsing}. The average length of texts in MR is 20 and that of SST is 19. We remove the neutral reviews and binarize labels for SST. We compose the ground truth dictionary $T$  by combing the MPQA subjective lexicon dataset \cite{Wilson2005Recognizing} and the sentiment lexicon dataset \cite{Hu2004Mining}. Only the strong subjective words in MPQA are considered as sentiment words. The ground truth dictionary $T$ contains 2006 positive words and 4783 negative words.

{\noindent \bf Results.}
We use 10-fold cross-validation to evaluate the performance of all methods except the TF-IDF. In each fold, we use the training dataset to train the models for sentiment classification and use the test dataset for sentiment word identification. Note that TF-IDF is a count-based feature selection method and  the TF-IDF values are fixed for a given dataset. Thus, we do not need to conduct cross validation for TF-IDF. We select the Top-k words with the highest TF-IDF values and compare with the ground truth.

Tables \ref{tb:accuracy} and \ref{tb:f1} show experimental comparisons of our methods, SV-AVG and SV-MAX, and three baselines in terms of the accuracy@k, precision, recall and F1 metrics. Tables \ref{tb:accuracy} first presents the sentiment classification accuracy. Results of the Accuracy@k are then shown in columns ``Top-k''.  Note that the TF-IDF and Saliency Map methods are feature selection methods which do not predict the class labels. Thus, we can not report the sentiment classification results for these two methods. We have the following observations:

(1). The SV-AVG method achieves the best performance on sentiment word identification, although its classification accuracy is slightly worse than the SV-MAX. Recall that the max pooling only keeps the highest value whereas the average pooling combines all values. Thus, the max pooling operation is good at finding the most discriminative features to separate texts into different classes because the classifier can predict the class accurately with the most discriminative features. On the contrary, the performance of SV-AVG for classification is not as good as SV-MAX because SV-AVG decreases the values of the most discriminative features by averaging all the feature values. In our sentiment word identification scenario, SV-MAX drops too much information by only keeping the maximum value, but SV-AVG with averaging the feature values keeps more useful information in the neural network. Therefore, SV-AVG is suitable for identifying all the discriminative words.

(2). The TF-IDF-softmax method performs better than the TF-IDF method, which indicates that supervised training the classifier can encode task-specific information in its parameters. Although the TF-IDF-softmax adopts the same classifier as our models, the performance of TF-IDF-softmax is worse than our methods. This demonstrates that  applying the convolutional operation on word embedding can capture more semantic information than the statistical features of TF-IDF.

(3). Although Saliency Map (SalMap-Max, SalMap-AVG) adopts the same CNN model, the performance of Saliency Map is only slightly better than the TF-IDF method and worse than our SV-AVG and SV-MAX. This indicates that the Saliency Map on word embedding captures much less information about identifying task-related words than the softmax parameters. We can further observe that the performance of SalMap-AVG is worse than SalMap-MAX. This is because
CNN with the max pooling operation can achieve better classification accuracy than using the mean pooling and the computed saliency map has larger gradient values to the discriminative features which are useful for identifying the task-specific words.

(4). We also notice that all methods generally achieve higher accuracy in SST than MR. This is because more texts in SST contain sentiment words than MR. It also indicates a positive correlation between sentiment word identification and classification accuracy.

We further compare the sentiment words selected by our models and baselines. Table \ref{tb:sent-words} shows the Top-10 most frequent words selected by each methods. The words are listed in order based on their frequency values in Top-5 lists. Comparing with the baselines, the Top-10 words selected by SV-AVG and SV-MAX are almost sentiment words; and the sentiment words are in the correct categories. It indicates the CNN model can figure out the most discriminative features from a sentence automatically. That is the reason why the CNN model achieves better performance for text classification than other methods, especially those methods which use the same classifier (i.e., TF-IDF-softmax).

\begin{table}[]
\centering
\caption{Top-10 sentiment words selected by our models and baselines}
\label{tb:sent-words}
\begin{adjustbox}{max width=0.95\textwidth}
\begin{tabular}{ccc}
\hline
               & Positive Words                                                                                                                          & Negative Words                                                                                                  \\ \hline
TF-IDF         & \begin{tabular}[c]{@{}c@{}}zone, good, ya, year, liked\\ like, fun, slight, worth, fantastic\end{tabular}                               & \begin{tabular}[c]{@{}c@{}}bad, movie, hate, time, year\\ good, work, just, characters, films\end{tabular}      \\ \hline
TF-IDF-Softmax & \begin{tabular}[c]{@{}c@{}}film, best, love, performances, good,\\ funny, fun, heart, work, performance\end{tabular}                    & \begin{tabular}[c]{@{}c@{}}movie, bad, just, like, feels,\\ plot, long, minutes, dull, thing\end{tabular}       \\ \hline
SalMap-MAX     & \begin{tabular}[c]{@{}c@{}}not, best, heart, performances, funny,\\ beautiful, beautifully, enjoyable, charming, beautiful\end{tabular} & \begin{tabular}[c]{@{}c@{}}bad, too, no, not, seems, lack, \\ worst, rather, instead, better, only\end{tabular} \\ \hline \hline
SV-MAX         & \begin{tabular}[c]{@{}c@{}}good, funny, best, well, love,\\ performances, fun, drama, great, family\end{tabular}                        & \begin{tabular}[c]{@{}c@{}}too, not, no, bad, only, \\ never, nothing, little, less, dull\end{tabular}          \\ \hline
SV-AVG         & \begin{tabular}[c]{@{}c@{}}best, live, great, entertaining, good,\\ fascinating, fun, beautifully, enjoyable, charming\end{tabular}     & \begin{tabular}[c]{@{}c@{}}bad, too, not, dull, no, boring\\ nothing, mess, problem, lack, bland\end{tabular}   \\ \hline
\end{tabular}
\end{adjustbox}
\end{table}

\subsection{Social Discrimination-related Word and Phrase Identification}

In this experiment, we aim to identify discrimination-related words and phrases (sexism-related words/ phrases or racism-related words/phrases) from tweets. These tweets are automatically labeled as about \textit{sexism}-related or \textit{racism}-related by hashtags. For example, if a tweet contains hashtag ``\#sexism'', we consider this tweet is related with sexism.

{\noindent \bf Datasets.}
We crawled tweets during the period from November 1, 2015 to April 17, 2016, which contain \textit{sexism} or \textit{racism} hashtags. We pre-process the tweets by tokenizing them and removing all punctuation and tokens beginning with the ``@'' symbol. We keep the tweets that contain more than 10 words. For those hashtags in tweets, we remove the hash signs and keep the words.

To evaluate the accuracy, we create a dataset $T_1$ containing 300 well-labeled tweets. In each tweet, the discrimination-related words are marked by two domain experts. The dataset is composed by 150 sexism-related tweets and 150 racism-related words. Meanwhile, to further conduct a case analysis about discrimination-related word and phrase identification, we compose another dataset $T_2$ which contains 10000 tweets with hashtag ``\#sexism'' and 10000 tweets with hashtag ``\#racism''.

{\noindent \bf Results of discrimination-related word identification.}
To evaluate the precision, recall and F1, we train all methods except the TF-IDF with 5 fold cross-validation on dataset $T_1$. Table \ref{tb:dwi} shows the precision, recall and F1 of discrimination-related words identification. In this experiment, we adopt the CNN model with the max pooling operation to compute the saliency map. The overall results are similar to the results of sentiment word identification. Our methods achieve the best performance on identifying the discrimination-related words.

\begin{table*}[]
\centering
\caption{Precision, recall and F1 on Discrimination-related Words Identification}
\label{tb:dwi}
\begin{adjustbox}{max width=0.95\textwidth}
\begin{tabular}{cccccccccc}
\hline
\multirow{2}{*}{Method} & \multicolumn{3}{c}{Top-1}                             & \multicolumn{3}{c}{Top-3}                             & \multicolumn{3}{c}{Top-5}                             \\ \cline{2-10}
                        & Precision        & Recall           & F1               & Precision        & Recall           & F1               & Precsion         & Recall           & F1               \\ \hline
TF-IDF                  & 11.33\%          & 4.07\%           & 5.99\%           & 8.44\%           & 9.10             & 8.76\%           & 7.80\%           & 14.01\%          & 10.02\%          \\ \hline
TF-IDF-Softmax          & 82.00\%          & 26.00\%          & 39.61\%          & 65.56\%          & 62.63\%          & 64.06\%          & 49.87\%          & 79.41\%          & 61.26\%          \\ \hline
SalMap-MAX 		        & 50.84\%          & 14.20\%          & 22.21\%          & 40.63\%          & 34.05\%          & 37.05\%          & 32.12\%          & 44.87\%          & 37.44\%          \\ \hline \hline
SV-MAX                  & 98.99\%          & \textbf{33.98\%} & 50.60\%          & 71.48\%          & 73.62\%          & 72.53\%          & \textbf{50.33\%} & \textbf{86.41\%} & \textbf{63.61\%} \\ \hline
SV-AVG                  & \textbf{99.65\%} & 33.92\%          & \textbf{50.62\%} & \textbf{72.80\%} & \textbf{74.34\%} & \textbf{73.56\%} & 50.14\%          & 85.34\%          & 63.17\%          \\ \hline
\end{tabular}
\end{adjustbox}
\end{table*}

Then, we conduct a detailed analysis about each method for task-specific word identification on dataset $T_2$. For each method, we build a dictionary of 100 words that occur most frequently in the Top-5 list of a tweet. We compare the dictionaries by calculating the overlapping size between the dictionary created by SV-AVG and that by each baseline. We find significant differences. Specifically, the ratios of overlapping words selected by SV-AVG and TF-IDF, TF-IDF-softmax, Saliency Map are 36\%, 58\% and 36\%, respectively.

Table \ref{tb:words} shows the Top-10 most frequent words selected by each methods. The words are listed in order based on their frequency values in Top-5 list. Comparing with the baselines, the Top-10 words selected by SV-AVG and SV-MAX contain more discrimination-related words. This is because SV-AVG and SV-MAX capture class information during the training process and filter out those words that are not related to the specific task.

Table \ref{tb:dis-words} shows several tweet examples where discrimination-related words selected by SV-AVG are highlighted. A word highlighted with the red color means the word in the Top-5 list. The superscript of a word indicates its ranking based on the score vector. We can see that our method can locate the discrimination-related words with high scores. Furthermore, our method can identify the words which are closely related to task-specific words such as ``girls'', ``\#white'', and ``\#Africa''.

\begin{table}[]
\large
\centering
\caption{Top-10 discrimination-related words selected by our models and baselines }
\label{tb:words}
\begin{adjustbox}{max width=0.95\textwidth}
\begin{tabular}{ccc}
\hline
               & Sexism-related Words                                                                                                      & Racism-related Words                                                                                                                   \\ \hline
TF-IDF         & \begin{tabular}[c]{@{}c@{}}live, miss, sexy, kikme, \\ cams, pornvideos, chat,\\  pussy, freeporn, cybersex\end{tabular}  & \begin{tabular}[c]{@{}c@{}}described, opposes, pbuh, prophet, \\ muhammad, christianity, selfie, \\ mecca, racism, racist\end{tabular} \\ \hline
TF-IDF-Softmax & \begin{tabular}[c]{@{}c@{}}sexism, women, men, feminism, \\ horny, online, sexy, \\ woman, pussy, kikme\end{tabular}      & \begin{tabular}[c]{@{}c@{}}racism, racist, ignorance, black, \\ pure,forms, opposes, \\ christianity, mecca, selfie\end{tabular}       \\ \hline
SalMap-MAX 	   & \begin{tabular}[c]{@{}c@{}}sexism, women, join, live, \\ online, cams, pussy, \\ pornvideos, kikme, freeporn\end{tabular} & \begin{tabular}[c]{@{}c@{}}racism, islam, racist, ignorance, \\ prophet, pbuh, end, \\ love, christianity, black\end{tabular}          \\ \hline \hline
SV-MAX         & \begin{tabular}[c]{@{}c@{}}sexism, women, men, woman,\\ feminism,sexist, female,\\ male, horny, sexy\end{tabular}         & \begin{tabular}[c]{@{}c@{}}racism, racist, black, white,\\ religion, hate, education,\\ america, isla, ignorance\end{tabular}          \\ \hline
SV-AVG         & \begin{tabular}[c]{@{}c@{}}sexism, women, men, feminism, \\  horny sexy, woman,\\  pussy, chat, cybersex\end{tabular}     & \begin{tabular}[c]{@{}c@{}}racism, racist, islam, black, \\ igonorance, opposes, religion,\\  mecca, white, hate\end{tabular}          \\ \hline
\end{tabular}
\end{adjustbox}
\end{table}

\begin{table}[]
\centering
\caption{Discrimination-related words selected by SV-AVG.}
\label{tb:dis-words}
\begin{adjustbox}{width=\textwidth}
\begin{tabular}{l}
\hline
\multicolumn{1}{c}{Racism-related Words}      \\ \hline
	 \tabitem \textcolor{red}{Idiots\textsuperscript{5}} like you can't even define \textcolor{red}{\#racism\textsuperscript{4}} or own your own but always calling \textcolor{red}{Black\textsuperscript{2}} people \textcolor{red}{racists\textsuperscript{1}} or \textcolor{red}{hateful\textsuperscript{3}}. \\

	 \tabitem \textcolor{red}{Hostility\textsuperscript{1}} between \textcolor{red}{minority\textsuperscript{2}} \textcolor{red}{races\textsuperscript{4}} always intrigue me. Who's the \textcolor{red}{enemy\textsuperscript{5}} here? Is there even one? \#369hong \#nonoboy \textcolor{red}{\#racism\textsuperscript{3}} \\

	 \tabitem Imagine a \textcolor{red}{\#white\textsuperscript{3}} person say \textcolor{red}{\#Africa\textsuperscript{1}} is too full of \textcolor{red}{brown\textsuperscript{2}} people or Arabia is too full of \#muslims? ... \textcolor{red}{\#racism\textsuperscript{5}} \textcolor{red}{\#whitegenocide\textsuperscript{4}} \\

	 \tabitem Cards against humanity is definitely \textcolor{red}{racist\textsuperscript{1}}, have 10 \textcolor{red}{'white'\textsuperscript{3}} cards. And then whoever is funniest gets a \textcolor{red}{'black'\textsuperscript{2}} \textcolor{red}{card\textsuperscript{5}} \textcolor{red}{\#racism\textsuperscript{4}} \\

	 \tabitem US cited for \textcolor{red}{police\textsuperscript{3}} \textcolor{red}{violence\textsuperscript{1}}, \textcolor{red}{\#racism\textsuperscript{2}} in scathing \#UN human \textcolor{red}{rights\textsuperscript{5}} review \#BlackLivesMatter \#policebrutality \textcolor{red}{\#torture\textsuperscript{4}} \\

	 \hline

\multicolumn{1}{c}{Sexism-related Words}   \\ \hline

	\tabitem The next time a \textcolor{red}{stranger\textsuperscript{2}} \textcolor{red}{man\textsuperscript{4}} asks me to smile, I'll punch him in the face! I \textcolor{red}{swear\textsuperscript{5}} it \textcolor{red}{\#Sexism\textsuperscript{1}} \textcolor{red}{\#Rude\textsuperscript{3}} \#StopSexualisingWomen \\

	\tabitem \textcolor{red}{Yeah\textsuperscript{5}} \textcolor{red}{fuck\textsuperscript{3}} \textcolor{red}{\#sexism\textsuperscript{2}} Unexpected \textcolor{red}{Job\textsuperscript{4}} Requirement Video Game Streamers Comfortable Wearing \textcolor{red}{Bikini\textsuperscript{1}} Top  \\

	\tabitem 80\% Telegraph readers: ok to call \textcolor{red}{boys\textsuperscript{5}} "sissies" to "man up" or \textcolor{red}{girls\textsuperscript{4}} studying \textcolor{red}{"male"\textsuperscript{3}} subjects \textcolor{red}{"lesbians"\textsuperscript{1}} \textcolor{red}{\#sexism\textsuperscript{2}}  \\

	\tabitem Shocking that many \textcolor{red}{\#women\textsuperscript{4}} are still experiencing \textcolor{red}{gender\textsuperscript{1}} \textcolor{red}{bias\textsuperscript{3}} at work and unacceptable levels of \textcolor{red}{\#sexism\textsuperscript{2}} in the \textcolor{red}{\#workplace\textsuperscript{5}} still exist \\

	\tabitem @EverydaySexism as a \textcolor{red}{female\textsuperscript{1}} Dr I'd \textcolor{red}{obviously\textsuperscript{5}} \textcolor{red}{wear\textsuperscript{4}} a mini \textcolor{red}{skirt\textsuperscript{3}} and have to be a 'baby' Dr \#juniordoctors \textcolor{red}{\#sexism\textsuperscript{2}}  \\ \hline

\end{tabular}
\end{adjustbox}
\end{table}

{\noindent \bf Results of discrimination-related phrase identification.}
For phrase identification, we set the phrase \textit{filter} $h=2$ and focus on identifying the discrimination-related phrases containing 2 words. Table \ref{tb:dis-phrases} shows several tweet examples where the Top-3 discrimination-related phrases are highlighted based on the score vector. Some phrases are longer than 2 words because the stop words are removed during the training phase. Our model successfully identifies the discrimination-related phrases such as ``black judge'', ``gender bias''. Meanwhile, the identified discrimination-related phrases usually contain the discrimination-related words such as ``sexism'', ``racism'', ``women'' and ``black''.

\begin{table}[]
\centering
\caption{Discrimination-related phrases selected by SV-AVG with filter $h=2$.}
\label{tb:dis-phrases}
\begin{adjustbox}{max width=0.95\textwidth}
\begin{tabular}{l}
\hline
\multicolumn{1}{c}{Racism-related Words} \\ \hline
\raisebox{-0.9\height}{ \includegraphics[width=0.9\textwidth]{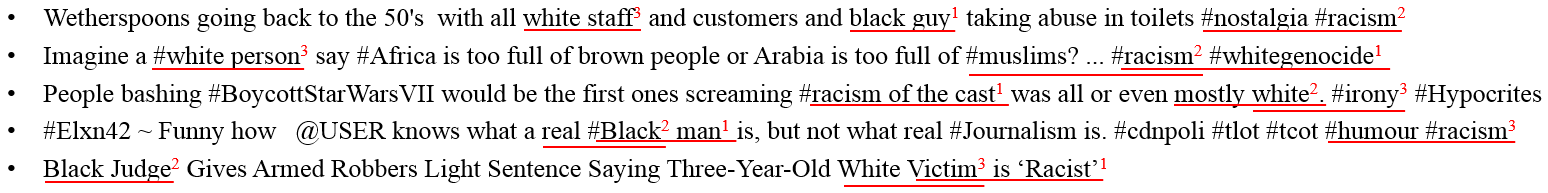} }     \\ \hline
\multicolumn{1}{c}{Sexism-related Words} \\ \hline
\raisebox{-0.9\height}{ \includegraphics[width=0.9\textwidth]{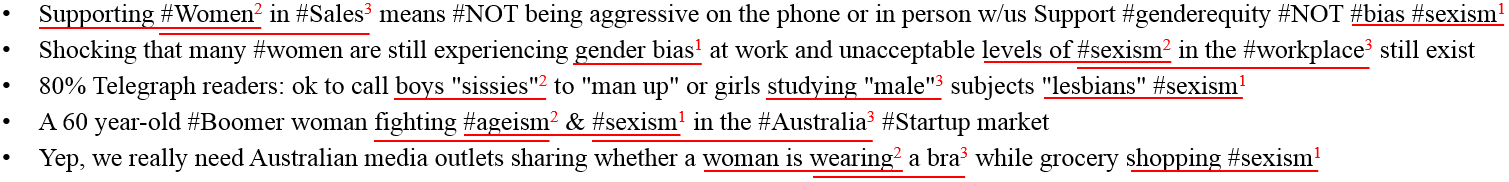} }      \\ \hline
\end{tabular}
\end{adjustbox}
\end{table}

\subsection{Fake Review Word and Phrase Identification}
In this experiment, we focus on identifying the words and phrases which are related to the fake review. The fake review detection aims to identify the reviews which try to mislead readers. There are a large number of websites (e.g., Amazon (\url{https://www.amazon.com}) and Yelp (\url{https://www.yelp.com})) which allow people to review the products or services which the users purchased. Many people rely on reviews for buying anything online. However, many reviews are generated by paid reviewers or rivals, which make the reviews unreliable. The fake review detection becomes a hot research topic recent years \cite{Ott2011Finding,Mukherjee2013Spotting,Li2014Towards}. In this work, we aim to figure out the fake review words and phrases, which can show the effectiveness of our model and further help to understand the pattern of fake reviews. Our model can also use to build blacklist words about the fake reviews, which can help to filter out the fake review automatically.

{\noindent \bf Datasets.}
We use the fake review corpus provided by \cite{Ott2011Finding,Ott2013Negative}, in which the truthful positive reviews were crawled from TripAdvisor and the deceptive positive reviews were submitted by Amazon Mechanical Turk. The reviews are about the most popular Chicago hotels. Each category of the corpus contains 400 reviews. In our experiment, we focus on identifying the truthful positive review words and deceptive positive review words.

{\noindent \bf Results of fake review word and phrase identification.}
Because there aren't ground-truth fake review words and phrases, in this experiment, we conduct a case analysis about fake review word and phrase identification by showing the words and phrases selected by our methods. We first show the Top-10 most frequent words from truthful and deceptive positive reviews selected by our method and baselines in Table \ref{tb:fr_compare}. Due to the space limit, we only report the result of SV-MAX. The words are listed in order based on their frequency values in Top-5 list.

We first focus on the words selected by SV-MAX. We can see that comparing with the first two words ``floor'' and ``bathroom'' in Top-10 truthful positive review words, the first words ``hotel'' and ``room'' in Top-10 deceptive positive review words are more general words to describe a hotel. Meanwhile, the rest of words in deceptive positive reviews contain highly polarized adjectives. Thus, we can figure out that the fake reviews prefer to use strong adjectives to describe the hotels. In contrast, the truth reviews have more nouns and moderate words related to the hotels. It also indicates that comparing with the true reviews, the CNN classifier assigns high weights to the adjectives to identify the fake reviews. We further compare the words selected by SV-MAX and baselines. In general, SV-MAX identifies more strong polarized words from deceptive positive reviews than the baselines, which indicates the effectiveness of our method.

For fake review phrase identification, we set the \textit{filter} $h=2$ and compare the phrases between truthful positive reviews and deceptive positive reviews selected by SV-MAX. Table \ref{tb:fr_phrases} shows one review example of each category where the Top-3 category-related phrases are highlighted based on the score vector. In general, our model can highlight the fake review phrases. Meanwhile, we can further see that although the truthful reviews also contain the sentiment words, the CNN classifier uses the nouns (i.e., locations) to identify the true reviews. It also means that the truthful reviews contain more specific description about the hotels.

\begin{table}[]
\centering
\caption{Top-10 words from truthful and deceptive positive reviews selected by our model and baselines}
\label{tb:fr_compare}
\begin{adjustbox}{max width=0.95\textwidth}
\begin{tabular}{ccc}
\hline
               & Deceptive Positive Words                                                                                                     & Truthful Positive Words                                                                                                   \\ \hline
TF-IDF         & \begin{tabular}[c]{@{}c@{}}hotel, room, hilton, \textbf{great}, rock,\\ really, east, family, hard, \textbf{loved}\end{tabular}                & \begin{tabular}[c]{@{}c@{}}hotel, great, really, large, service, \\ suite, room, free, place, talbott\end{tabular}        \\ \hline
TF-IDF-Softmax & \begin{tabular}[c]{@{}c@{}}chicago, husband, visit, amazing, family,\\ staying, looking, spa, vacation, \textbf{luxury}\end{tabular}  & \begin{tabular}[c]{@{}c@{}}floor, bathroom, location, large, small,\\ rate, street, reviews, michigan, great\end{tabular} \\ \hline
SalMap-MAX 	   & \begin{tabular}[c]{@{}c@{}}experience, family, spa, husband, \textbf{luxury},\\ \textbf{luxurious}, vacation, wedding, wife, food\end{tabular} & \begin{tabular}[c]{@{}c@{}}floor, large, small, reviews, rate, \\ street, priceline, blocks, wife, upgraded\end{tabular}  \\ \hline \hline
SV-MAX         & \begin{tabular}[c]{@{}c@{}}hotel, room, \textbf{great}, \textbf{comfortable}, \textbf{like}, \\ recommend, \textbf{amazing}, \textbf{luxury}, business, \textbf{clean}\end{tabular} & \begin{tabular}[c]{@{}c@{}}floor, bathroom, large, small, location\\ street, blocks, booked, river, parking\end{tabular}  \\ \hline
\end{tabular}
\end{adjustbox}
\end{table}

\begin{table}[]
\centering
\caption{Truthful and deceptive review phrases selected by SV-MAX with filter $h=2$.}
\label{tb:fr_phrases}
\begin{adjustbox}{max width=0.98\textwidth}
\begin{tabular}{c m{0.88\textwidth}}
\hline
Truthful Positive Reviews  & \raisebox{-0.9\height}{ \includegraphics[width=0.85\textwidth, ]{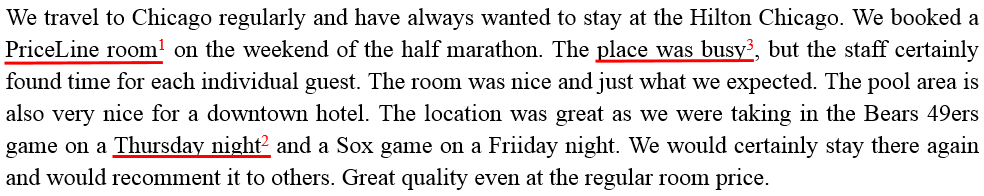} }   \\ \hline
Deceptive Positive Reviews & \raisebox{-0.92\height}{ \includegraphics[width=0.85\textwidth, ]{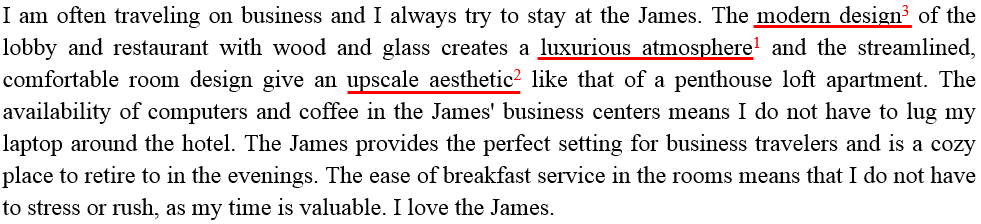} } \\ \hline
\end{tabular}
\end{adjustbox}
\end{table}

\section{Conclusion}
In this paper, we have focused on task-specific word or phrase identification based on the convolutional neural network. We show how to derive the hidden score vector from CNN parameters and why the score vector can be used to identify task-specific words or phrases. Experimental results on sentiment word identification showed that our approach can significantly improve the accuracy for identifying the task-specific words compared with state-of-the-art methods including TF-IDF, softmax classifier and saliency map. We further showed that our approach can successfully identify the task-specific words or phrases effectively from discrimination-related tweets and fake review dataset. In our future work, we will extend our approach to identify key sentences from a document by using the convolutional document model.

\section*{Acknowledgements}
The authors acknowledge the support from the National Natural Science Foundation of China (71571136), the 973 Program of China (2014CB340404), and the Research Projects of Science and Technology Commission of Shanghai Municipality (16JC1403000, 14511108002) to Shuhan Yuan and Yang Xiang, and from National Science Foundation (1646654,1564250) to Xintao Wu.

\clearpage

\end{document}